\pdfoutput=1

\documentclass[11pt]{article}

\usepackage[]{acl}

\usepackage{times}
\usepackage{booktabs}
\usepackage{latexsym}
\usepackage{tabularray}
\usepackage{graphicx}
\usepackage{multirow}
\usepackage{tabularx}
\usepackage{booktabs}
\usepackage{makecell}
\usepackage{adjustbox}
\usepackage{float}
\usepackage{amsmath}
\usepackage{subcaption}

\usepackage[T1]{fontenc}

\usepackage[utf8]{inputenc}

\usepackage{microtype}

\usepackage{inconsolata}

\title{ICLEF: In-Context Learning with Expert Feedback for Explainable Style Transfer}

\author{Arkadiy Saakyan$^1$ 
\and Smaranda Muresan$^{1,2}$ 
\\
$^1$ Department of Computer Science, Columbia University \\
$^2$ Data Science Institute, Columbia University\\
\texttt{a.saakyan@cs.columbia.edu, smara@columbia.edu} \\ 
}

\begin{document}
\maketitle
\begin{abstract}
While state-of-the-art large language models (LLMs) can excel at adapting text from one style to another, current work does not address the \textbf{explainability} of style transfer models. Recent work has explored generating textual explanations from larger teacher models and distilling them into smaller student models. One challenge with such approach is that LLM outputs may contain errors that require expertise to correct, but gathering and incorporating expert feedback is difficult due to cost and availability. To address this challenge, we propose \texttt{ICLEF}, a novel human-AI collaboration approach to model distillation that incorporates \textbf{scarce} expert human feedback by combining \textit{in-context learning} and \textit{model self-critique}. We show that our method leads to generation of high-quality synthetic explainable style transfer datasets for formality ({\textsc{e-GYAFC}}) and subjective bias ({\textsc{e-WNC}}). Via automatic and human evaluation, we show that specialized student models fine-tuned on our datasets outperform generalist 
teacher models on the explainable style transfer task in one-shot settings, and perform competitively compared to few-shot teacher models, highlighting the quality of the data and the role of expert feedback. In an extrinsic task of authorship attribution, we show that explanations generated by smaller models fine-tuned on {\textsc{e-GYAFC}} are more predictive of authorship than explanations generated by few-shot teacher models.
\end{abstract}
\section{Introduction}

\begin{figure*}[ht]
    \centering
    \includegraphics[width=5in]{ 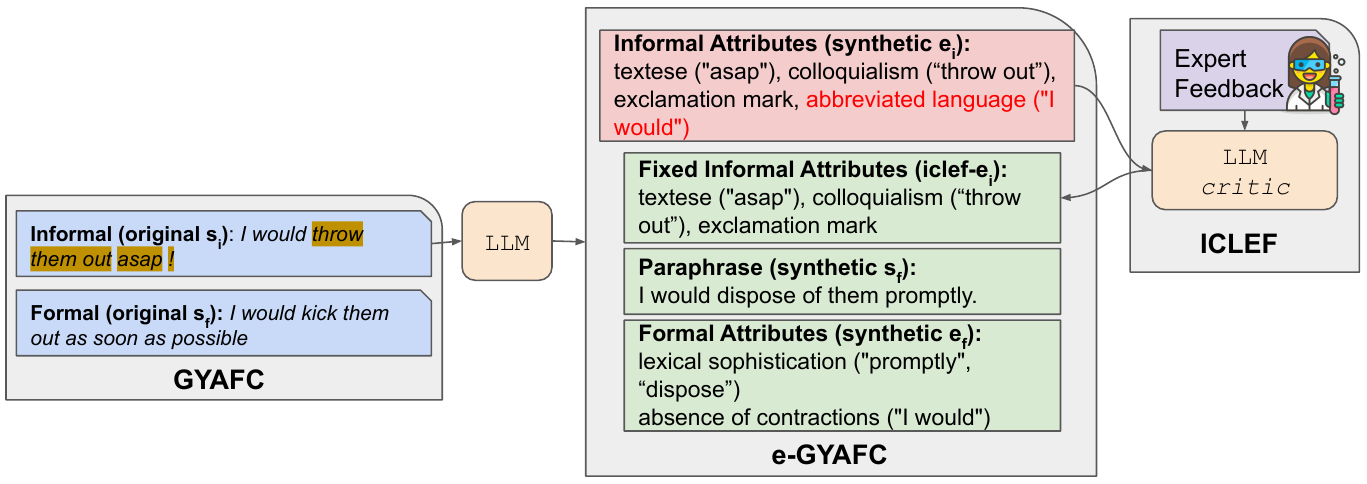}
    \caption{Generating \textsc{e-GYAFC}: formality style transfer dataset GYAFC \cite{rao-tetreault-2018-dear} is augmented with semi-structured natural language explanations. The LLM generates the informal attributes of the input sentence, a formal paraphrase, and the formal attributes of the resulting sentence. Expert feedback is incorporated via in-context learning and self-critique to refine the initial generations.}
    \label{fig:intro_fig}
\end{figure*}

Attribute style transfer is the task of transforming a given text along a particular style dimension, such as changing its formality, bias, or level of offensiveness \cite{lample2018multipleattribute, sudhakar-etal-2019-transforming, jin-etal-2022-deep}. Formality style transfer (e.g., informal$\rightarrow$formal) could be useful in any writing assistance system, while neutralizing text that contains subjective bias would be an important tool for Wikipedia editors \cite{wnccorpus} or journalists \cite{rosenberg_fischer2023newsrooms}. 

Style transfer approaches have primarily focused on the text re-writing task (e.g., informal $\rightarrow$formal,  subjective bias $\rightarrow$ neutral) using various methods from supervised  \cite{rao-tetreault-2018-dear, wnccorpus,zhong-etal-2021-wikibias-detecting} to unsupervised \cite{krishna-etal-2020-reformulating} and zero-shot methods using LLMs \cite{reif-etal-2022-recipe} (see also \citet{jin-etal-2022-deep} for a survey on style transfer). However, to our knowledge, no effort has focused on providing \emph{textual explanations} for the style transfer task. For example, when transforming an informal sentence ``I would throw them out asap !'' into a formal paraphrase ``I would dispose of them promptly'' it would be useful to provide an explanation of the informal attributes in the input sentence (e.g., textese (``asap"), colloquialism (``throw out")), and formal attributes for the paraphrase (e.g., lexical sophistication (``promptly" and ``dispose"); lack of abbreviations (``I would")). Similarly, for neutralizing subjective bias in ``Orbis latinus, integral site of romance language" $\rightarrow$ ``Orbis latinus, comprehensive site of romance language", it would be useful to have an explanation about which word/phrase in the input is biased and why as well as the type of bias (e.g., Framing (``integral" implies a subjective evaluation on the site's importance)).        
The model's explanations could help the user better assess the correctness of the style transfer system, could be used as features in downstream tasks such as authorship attribution (Section \ref{sec:apps}), or could act as a defense against spurious correlations \cite{ludan2023explanationbased, NEURIPS2018_4c7a167b} and annotation artifacts \cite{mccoy-etal-2019-right, poliak-etal-2018-hypothesis}.

\begin{figure*}[ht]
    \centering
    \includegraphics[width=5in]{ 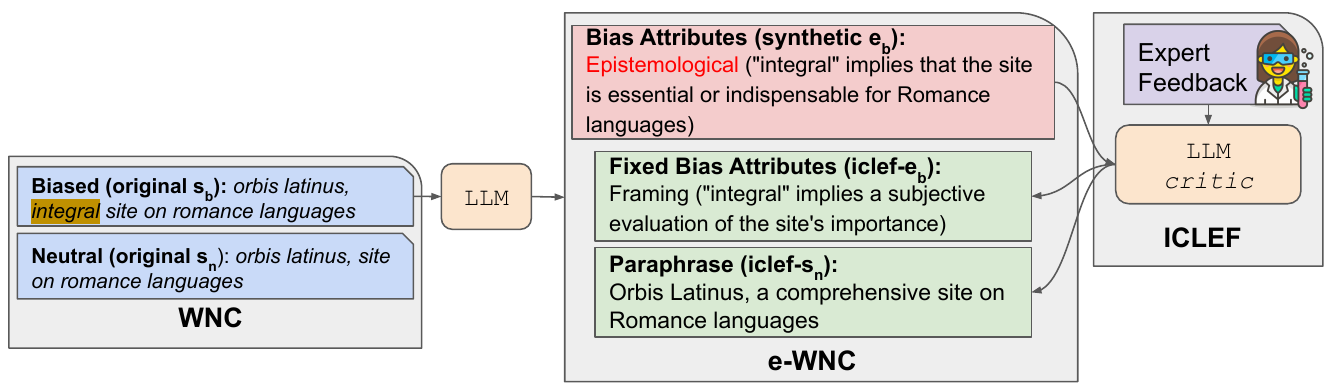}
    \caption{Generating \textsc{e-WNC}: WNC \cite{wnccorpus} is augmented with natural language explanations. The LLM generates the bias attributes of the input sentence and an unbiased paraphrase. Expert feedback is incorporated via in-context learning and self-critique to refine the initial generations.}
    \label{fig:intro_fig_bias}
\end{figure*}

To enable explainability in style transfer models, we provide the following contributions:
\begin{itemize}
    \item A new task of \textit{explainable} style transfer for which, in addition to sentence rewriting, 
    the model needs to generate textual explanations.
    
    \item A novel human-AI collaboration framework, \textit{In Context-Learning with Expert Feedback \texttt(ICLEF)} (see Figure \ref{fig:intro_fig}, Figure \ref{fig:intro_fig_bias},  and \S \ref{sec:iclef}). The approach combines model distillation for explanation generation \cite[]{Ho2022LargeLM, magister2023teaching} with self-critique ability of LLMs \cite[][\textit{inter alia}]{madaan2023selfrefine, bai2022constitutional, saunders2022selfcritiquing, scheurer2023training}, where the critic, unlike in prior work, is instantiated with expert demonstrations.

    \item Using ICLEF, we create for the first time
    datasets for explainable style transfer by augmenting an existing formality style transfer dataset GYAFC \cite{rao-tetreault-2018-dear} and the neutralizing subjective bias dataset WNC \cite{wnccorpus} with textual explanations (\S \ref{sec:egyafc}). We show that the datasets generated with the help of ICLEF, \textsc{\textsc{e-GYAFC}} and \textsc{\textsc{e-WNC}}, are of
    good quality via automatic and expert evaluation, and that ICLEF-fixed instances are preferred (\S \ref{section:dataQuality}). 

    \item Experiments that show that 
    student models outperform teacher models in one-shot setting and perform comparably even with few-shot teacher models in automatic and expert evaluation, confirming the utility and quality of the datasets (\S \ref{sec:exps}). Moreover, in an extrinsic evaluation, we show that explanations generated by student models fine-tuned on our data produce a better signal for the authorship attribution task than the explanations produced by few-shot teacher models (\S \ref{sec:apps}). 

\end{itemize}
    We release the data, models, and code to encourage further research on explainability, learning from scarce human feedback, and style transfer. 
\footnote{\href{https://github.com/asaakyan/explain-st}{github.com/asaakyan/explain-st}}

\section{Related Work}

\paragraph{Knowledge distillation and human feedback}
Model or knowledge distillation is a process of fine-tuning a smaller student model to imitate the behaviour of a more competent teacher model \cite{Beyer_2022_CVPR, 10.1145/1150402.1150464, hinton2015distilling}. Knowledge distillation became a popular technique, allowing to generate datasets of similar quality to the crowd-sourced data \cite{west-etal-2022-symbolic}, especially when combined with a model-in-the-loop approach \cite{wiegreffe-etal-2022-reframing, bartolo-etal-2022-models, chakrabarty-etal-2022-flute}. Recent work explores model distillation with natural language explanations \cite{wang2023pinto, Ho2022LargeLM, magister2023teaching}, showing that large language models are capable of generating acceptable enough reasoning steps for student models to learn.
Approaches to incorporate human feedback such as RLHF \cite{NEURIPS2020_1f89885d} and DPO \cite{rafailov2023direct} require large amounts of crowdsourced data and have not been generally shown to be effective for expert preferences. Imitation learning from human feedback (ILF) \cite{scheurer2023training} utilizes human feedback to improve model-generated instances, and then fine-tunes on that data. Unlike these works, we focus on incorporating \textit{expert} feedback which is naturally scarce and expensive to collect.
Unlike other self-critique approaches \cite[]{madaan2023selfrefine, bai2022constitutional, saunders2022selfcritiquing}, we condition the model on expert corrections to incorporate high-quality human feedback.

\paragraph{Textual Explanations}
Natural language explanations have been utilized for a variety of tasks \citet{NEURIPS_DATASETS_AND_BENCHMARKS2021_698d51a1}, such as natural language inference \cite{NEURIPS2018_4c7a167b}, commonsense \cite{rajani-etal-2019-explain, aggarwal-etal-2021-explanations}, social norm entailment \cite{chwang2023sociocultural}. We focus on creating natural language explanations for the style transfer task, which has not been addressed before. 

\paragraph{Style transfer} Style transfer approaches range from instruction-based methods \cite{reif-etal-2022-recipe} and paraphrasing \cite{krishna-etal-2020-reformulating}, to approaches focused on learning in low-resource settings \cite{patel2022lowresource}. Much of style transfer work focuses on style representations that decouple style and content \cite{wegmann-etal-2022-author, wegmann-nguyen-2021-capture}, however most of these methods are not designed to be interpretable. Interpretable approaches rely on constructing interpretable embeddings, such as LIWC \cite{tausczik2010psychological} or LISA \cite{patel2023learning}. \citet{zhong-etal-2021-wikibias-detecting} proposed identifying biased segments in conjunction with neutralizing biased text. Unlike these approaches, we propose to use natural language explanations to further enhance model interpretability.

\section{Building Datasets for Explainable Style Transfer} \label{sec:egyafc}
We build two explainable style transfer datasets by first augmenting existing datasets with synthetic textual explanations generated by a teacher model (\S \ref{sec:augmenting}), and then improving the generated data using our In-Context Learning with Expert Feedback (\texttt{ICLEF}) framework (\S \ref{sec:iclef}).

\subsection{Augmenting Style Transfer Datasets with Synthetic Textual Explanations} \label{sec:augmenting} 
\paragraph{Formality style transfer} The GYAFC \cite{rao-tetreault-2018-dear} formality style transfer dataset contains parallel formal and informal sentences. The informal sentences are collected from Yahoo Answers, and formal paraphrases were crowdsourced using Amazon Mechanical Turk (AMT). 
We use ChatGPT-3.5 to generate explanations and formulate the following multi-step generation task: given an informal sentence from GYAFC $s_i$, generate a structured explanation of its informal attributes $e_i$, then generate a formal paraphrase $s_f$ based on these attributes, then the formal attributes of the resulting paraphrase $e_f$. Generating both $e_i$ and $e_f$ allows us to train models in both directions (formal $\rightarrow$ informal and informal $\rightarrow$ formal). 
We use a semi-structured format for the explanations. Specifically, we ask the model to generate a list of attributes followed by an excerpt from the sentence as the evidence: attribute (``evidence''), see examples in Figure \ref{fig:intro_fig}. These explanations have a consistent format, making it easier to verify and automatically evaluate. 

\paragraph{Subjective bias style transfer} We focus on the task of neutralizing subjective biased language introduced by \citet{wnccorpus} to make sentences follow the 
Wikipeida Neutral Point of View Policy.\footnote{\href{https://simple.wikipedia.org/wiki/Wikipedia:Neutral_point_of_view}{Wikipedia.org}} We start with the Wikipedia Neutrality Corpus (WNC) \cite{wnccorpus}, a parallel corpus of 180,000 sentence pairs originating from
Wikipedia edits of subjective biased language. The goal is to generate an explanation ($e_b$) for the type of bias present in the biased sentence ($s_b$), following the scheme proposed by \citet{wnccorpus} and \citet{recasens-etal-2013-linguistic}: Framing, Epistemological, and Demographic (see definitions in Appendix \ref{app:biasStats}). It is estimated by \citet{wnccorpus} that a small percentage of cases, the instances in the WNC contain noise. We add an additional ``No Bias'' label for these cases to reduce hallucinated biases for neutral sentences. In this case, we ask the model to output ``This sentence does not contain bias'' as the explanation (see second WNC example in Table \ref{tab:iclefVsNo}). The explanation is structured as Type of Bias ("evidence" reasoning). Then, the teacher model generates an unbiased paraphrase ($s_n$). See Figure \ref{fig:intro_fig_bias} for an overview. 
At the time we developed this dataset, ChatGPT-4 became available, so we use this more powerful model as a teacher, especially since generating explanations for this task might require more reasoning capabilities.
We do not generate explanations for the neutrality of the paraphrase as we are not 
exploring the neutral to biased paraphrase direction due to ethical concerns.

\subsection{In-Context Learning from Expert Feedback (\texttt{ICLEF})} \label{sec:iclef}
ChatGPT generations might contain errors (e.g., the generated style attribute "abbreviated language" with the evidence ``I would'' in Figure \ref{fig:intro_fig}).
To improve the quality of the data, we turn to expert feedback, since previous work has identified that crowdworkers on platforms such as Amazon Mechanical Turk could be unreliable  for open-ended generation tasks \cite{karpinska-etal-2021-perils}, and 
might even rely on ChatGPT to provide their answers \cite{veselovsky2023artificial}. The crux of our approach is to combine in-context learning and self-critique abilities of LLMs by instantiating the LLM-critic model with few-shot expert feedback demonstrations.

\paragraph{\textsc{e-GYAFC}} 
For the formality style transfer task, we hire an expert annotator with a Masters degree in linguistics on Upwork\footnote{\href{https://www.upwork.com/}{Upwork.com}}.
Our annotation protocol (see Appendix \ref{app:protocols}) provides a non-exhaustive reference to formality and informality attributes and asks the annotator to provide feedback on which attributes in $e_i, e_f$ are incorrect if any, among other information. We provide 50 random samples for annotation. The annotation process took each expert only 2-3 hours. We find that the rate of critical errors observed in formality explanations is significantly lower ( ($\approx$ 8\% for formality explanations vs $\approx$ 56\% for informality explanations), so we only focus on applying the LLM-critic to incorrect informality attributes ($e_i$).

To do so, we instantiate an LLM-Critic model by prompting another LLM (ChatGPT-3.5) with 35 expert human feedback corrections in-context (we find that this number leads to satisfactory correctness of $\approx$ 87\%, see Appendix \ref{app:iclefNumFeed} for how performance changes given the amount of feedback) and ask it to act as an annotator on the new instances to identify incorrect attributes in them (see prompt in Table \ref{tab:prompts} in Appendix \ref{app:prompts}). To mitigate the risk of generating new incorrect attributes, we only query the model to identify and remove incorrect attributes in $e_i$, and if there are any, provide a fixed \textit{iclef}-$e_i$ where they are removed. We refer to the resulting model as LLM-Critic (see how \textit{abbreviated langauge} is removed in  Figure \ref{fig:intro_fig}).

In this way, we fix $\approx 30\%$ of the generated data (2853 instances) -- all instances for which the critic has predicted that an improvement is needed.
The resulting data (which we refer to as \textsc{\textsc{e-GYAFC}}) contains 9,960 original $s_i$, synthetic $s_f$ instances with corresponding  \textit{iclef}-$e_i$, synthetic $e_f$ attribute explanations. We randomly split the data into 8,000 training instances and 1,960 held-out test instances.

\paragraph{\textsc{e-WNC}} For the neutralizing subjective bias task, we hire an expert annotator with a PhD in linguistics on Upwork. We provide 50 random samples ensuring equal representation for each type of bias. To counteract potential annotation biases, we decided to use two annotators, one of the authors acting as the additional annotator, and then randomly sample the instances in equal proportions.
The annotation protocol asks to provide a corrected explanation instead of $e_b$ if the synthetic explanation contains an incorrect type of bias or wrong justification. We randomly sample 35 distinct instances of the two annotators' feedback, finding that this number leads to satisfactory correctness of $\approx$ 93\% (see Appendix \ref{app:iclefNumFeed}). We provide the expert critiques in-context in a similar way to \textsc{\textsc{e-GYAFC}} (see bottom prompt in Appendix \ref{app:prompts}, Table \ref{tab:prompts}). Since new bias attributes might have been introduced, we regenerate the paraphrase $s_n$ given the new explanation (see corrected type of bias and a new paraphrase in Figure \ref{fig:intro_fig_bias}). 
We fix 8\% of synthetic explanations in this manner (the lower rate of errors is explained by the higher quality of initial generations due to the use of a more powerful ChatGPT-4 model). The resulting dataset (\textsc{\textsc{e-WNC}}) contains 3,000 original WNC $s_b$ biased sentences, \textit{iclef}-${e_b}$ bias explanations, as well as the corresponding \textit{iclef}-$s_n$ neutralized sentences. We randomly split the data into 2,500 training instances and 500 held-out test instances.
 
\begin{table*}[h]
\centering
\small
\begin{tabularx}{\textwidth}{|X|X|X|}
\hline
\textbf{Informal} ($s_{i}$) & \textbf{Gen. expl.} (synthetic $e_i$) & \textbf{ICLEF expl.} (\textit{iclef}-$e_i$) \\ 
\hline
hopefully you aren't too old or you are screwed. & \textcolor{red}{informal greeting ("hopefully")}, slang ("screwed"), contraction ("aren't") & slang ("screwed"), contraction ("aren't") \\ 
\hline
more info, we are both in our very late twenties. & [...], \textcolor{red}{omission of prepositions ("in our very late twenties")} & abbreviation ("info"), colloquialism ("very late twenties") \\ 
\hline\hline
\textbf{Biased} ($s_{b}$) & \textbf{Gen. expl.} (synthetic $e_b$) & \textbf{ICLEF expl.} (\textit{iclef}-$e_b$) \\ 
\hline
[...] a play on the title of the popular mtv series, "unplugged". & \textcolor{red}{Epistemological ("popular" implies that the MTV series is universally well-liked)} & Framing ("popular" is a subjective term that implies the MTV series is widely liked) \\ 
\hline
[..] kendal, cbe (born 25 september 1946) is an english actress known in the united kingdom [...]. & \textcolor{red}{Demographic ("actress" implies that the person is female and could perpetuate gender stereotypes or assumptions)} & This sentence does not contain bias. \\ \hline
claims for the existence of paranormal psychic abilities such as clairvoyance are highly controversial. & \textcolor{red}{This sentence does not contain bias.} & Epistemological ("highly controversial" implies that the existence of paranormal psychic abilities is widely disputed) \\ 
\hline
\end{tabularx}
\caption{Qualitative comparison of dataset instances before and after application of ICLEF.}
\label{tab:iclefVsNo}
\end{table*}

\subsection{Dataset quality} \label{section:dataQuality}
\paragraph{Automatic evaluation of paraphrase quality} We estimate paraphrase quality automatically using Mutual Implication Score (MIS) \cite{babakov-etal-2022-large} and Formality Score
(see \S \ref{subsec:metrics} for metrics details) between our formal paraphrases and the ones in GYAFC.
We find that our paraphrases are of comparable quality with an MIS of 81.30 vs. 83.08 for GYAFC, yet we achieve a higher formality score of 98.43 vs. 89.39 for GYAFC (see Table \ref{tab:autoevalPara}). For example, for the GYAFC example in Figure \ref{fig:intro_fig}, the formal paraphrase  contains \textit{kick them out}). 

Similarly, for the \textsc{e-WNC} dataset we report bias score from an off-the-shelf classifier (see \S \ref{subsec:metrics}) along with MIS.
The MIS scores are 79.32 for original paraphrases vs. 85.58 for our paraphrases, indicating higher semantic similarity. The neutrality scores are 69.34 vs. 72.64, indicating higher neutrality of our paraphrases (see Table \ref{tab:autoevalPara}).

\begin{table}[h]
\centering
\small
\begin{tabular}{lcccc}
\toprule
& \multicolumn{2}{c}{\textbf{e-GYAFC}} & \multicolumn{2}{c}{\textbf{e-WNC}} \\
\cmidrule(lr){2-3} \cmidrule(lr){4-5}
& \textbf{MIS } & \textbf{Formality } & \textbf{MIS } & \textbf{Neutrality } \\
\midrule
Orig. para. & \textbf{83.08} & 89.39 & 79.32 & 69.34 \\
Cand. para. & 81.30 & \textbf{98.43} & \textbf{85.58} & \textbf{72.64} \\
\bottomrule
\end{tabular}
\caption{\label{tab:autoevalPara} Synthetic paraphrases (generated via model distillation for \textsc{e-GYAFC} and \textsc{e-WNC}) exhibit higher quality overall in automatic evaluation compared to original paraphrases (from GYAFC and WNC, respectively).}
\end{table}

\begin{figure}[h]
    \centering
    \includegraphics[width=0.6\columnwidth]{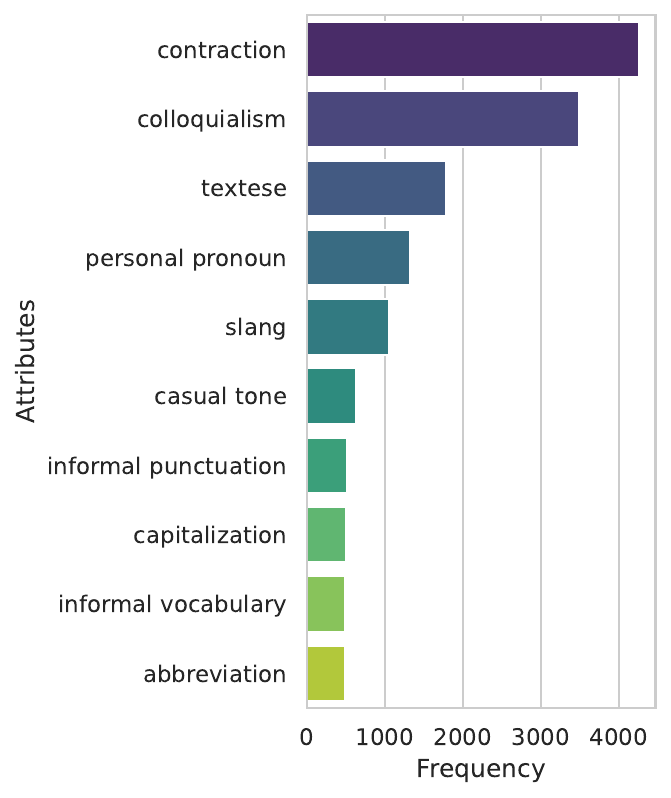}
    \caption{Top 10 informal attributes. See top 50 (in)formality attributes in Appendix Figure \ref{fig:informalDistr}, \ref{fig:formalDistr}).}
    \label{fig:top10inform}
\end{figure}

\paragraph{Human evaluation}
For \textsc{e-GYAFC} we hire 3 expert annotators, 2 of which performed the annotation, as well as an independent expert annotator with a masters degree in linguistics.
We ask their preferences on 100 randomly sampled instances with respect to the explanations (synthetic $e_i$ vs. \textit{iclef}-$e_i$) and paraphrases (original $s_f$ in GYAFC vs. synthetic $s_f$ in \textsc{e-GYAFC}). In addition, we ask for acceptability judgments (whether the paraphrase or explanation are correct and complete) for the preferred paraphrase and separately for $e_f$. We report preference or equal rates and acceptability rates.\footnote{We compute preference or equal preference among acceptable instances. For acceptability, we compute dispreferred instances as unacceptable.}
Overall, we found that our dataset instances are considered acceptable, with the average acceptability rate for $e_i, s_f, e_f$ being 87\%, 77\%, 98\% respectively (row 1 in Table \ref{tab:humaneval}). Synthetic paraphrases are generally preferred to the ones in the GYAFC corpus (average 77\%), and  \textit{iclef}-explanations are preferred or are equal in quality with original generations on average in 90\% of cases (row 2 in Table \ref{tab:humaneval}).  We computed the pairwise accuracy between annotator responses for all categories of \textsc{e-GYAFC} evaluation, and found that it averages at 81\% across all categories. We provide more details in Appendix \ref{app:human}. Table \ref{tab:iclefVsNo} shows qualitative examples of successful edits with \texttt{ICLEF}. Figure \ref{fig:top10inform} shows the top 10 most frequent informality attributes. 

\begin{table}[h]
\small
\centering
\label{tab:comparison_rates}
\begin{tabular}{@{}lcccccc@{}}
\toprule
& \multicolumn{3}{c}{\textbf{e-GYAFC}} & \multicolumn{2}{c}{\textbf{e-WNC}} \\
\cmidrule(lr){2-4} \cmidrule(lr){5-6}
& \textbf{e$_i$} & \textbf{s$_f$} & \textbf{e$_f$} & \textbf{e$_b$} & \textbf{s$_n$} \\
\midrule
\textbf{Acceptability} & 87\% & 77\% & 98\% & 73\% & 74\% \\
\textbf{Preference} & 90\% & 77\% & - & 78\% & 77\% \\
\bottomrule
\end{tabular}
\caption{\label{tab:humaneval} Acceptability and Preference Rates (between synthetic explanation vs.  \textit{iclef} explanation, and synthetic paraphrase vs. original paraphrase form the dataset) for \textsc{e-GYAFC} and \textsc{e-WNC}.
}
\end{table}

For \textsc{e-WNC}, we hire 2 annotators, one of whom performed the \texttt{ICLEF} annotation. To meaningfully evaluate the preference of \textit{iclef}-explanations compared to the synthetic ones, we ask to provide feedback on 50 instances where the explanation has been updated. The average preference for  \textit{iclef}-$e_b$ compared to synthetic $e_b$ is 78\%. Average acceptability rate of \textit{iclef}-$e_b$ is 73\%. The average preference for synthetic $s_n$ compared to the neutral sentence in the \textsc{e-WNC} corpus is 76\%. Average acceptability rate of synthetic $s_n$ is 74\%. The pairwise accuracy between the annotators is 77\%. More details can be found in Appendix \ref{app:humanBias}. 

\begin{table*}[h!]
\begin{adjustbox}{width=\textwidth,center}
\centering
\small
\begin{tabular}{lccccccccccc}
\toprule
 & & \multicolumn{4}{c}{Formal $\rightarrow$ Informal} & \multicolumn{4}{c}{Informal $\rightarrow$ Formal} & \\
\cmidrule(lr){3-6} \cmidrule(lr){7-10}
Model & Size & \makecell{Form.Attrs. \\ BLEU} & \makecell{MIS} & \makecell{Informality} & \makecell{Inform.Attrs. \\ BLEU} & \makecell{Inform.Attrs. \\ BLEU} & \makecell{MIS} & \makecell{Formality} & \makecell{Form.Attrs. \\ BLEU} & \textbf{Average} \\
\midrule
\makecell[l]{Vic$_1$} & 13B & 23.16 & 83.24 & 35.26 & 10.97 & 27.31 & 61.22 & \underline{\textbf{98.70}} & 9.88 & 43.72\\
\makecell[l]{Vic$_5$} & 13B & 24.16 & 85.18 & 33.58 & 13.09 & 27.95 & 73.18 & 98.20 & 15.45 & 46.35\\
\makecell[l]{Vic$_{10}$} & 13B & 19.78 & 82.00 & 49.17 & 12.35 &30.97 & 73.26 & 97.86 & 17.14 & 47.82 \\
\makecell[l]{GPT-3.5$_1$} & ? & 28.88 & 90.12 & 39.65 & 12.80 & 27.65 & 81.98 & 97.68 & 10.36 & 48.64\\
\makecell[l]{GPT-3.5$_5$} & ? & 33.98 & \underline{\textbf{90.37}} & 38.23 & 16.14 & 36.03 & 85.37 & 97.54 & 16.30 & 51.74\\
\makecell[l]{GPT-3.5$_{10}$} & ? & \underline{36.78} & 89.57 & \underline{49.81} & \underline{18.51} & \underline{37.36} & \underline{\textbf{85.62}} & 97.75 & \underline{20.31} & \underline{54.46} \\
\midrule \midrule
\makecell[l]{LLaMA$_{\rightarrow}$} & 7B & 39.64 & 85.31 & 61.33 & 19.86 & 38.02 & 81.80 & 97.77 & 25.10 & 56.10\\
\makecell[l]{Alpaca$_{\leftrightarrow}$} & 7B & \textbf{40.42} & 81.76 & \textbf{66.71} & \textbf{21.11} & \textbf{40.34} & 79.43 & 98.57 & \textbf{25.75} & \textbf{56.76}\\
\bottomrule
\end{tabular}
\end{adjustbox}
\caption{Performance of instruction-tuned and fine-tuned models on the explainable formality style transfer task. Best bolded, best non-fine-tuned underlined.}
\label{tab:instruct}
\end{table*}

\section{Evaluation of Student and Teacher Models on the Explainable Style Transfer Task} \label{sec:exps}
We focus on evaluating the performance of select large language models on the \textit{explainable} style transfer task. 
We do not evaluate post-hoc rationale systems (generating attributes given the paraphrase pair), since such pipeline models are less likely to reflect the underlying reasons for the model prediction, while models that jointly predict and rationalize exhibit desirable properties for faithful explanations \cite{wiegreffe-etal-2021-measuring}.
For explainable formality style transfer, 
we test the generation of $e_f, s_i, e_i$ given $s_f$ (Formal$\rightarrow$Informal) and $e_i, s_f, e_f$ given $s_i$ (Informal$\rightarrow$Formal) on a held-out test set from \textsc{e-GYAFC}. We evaluate how closely the model generated $e_i, e_f$ match \textsc{e-GYAFC} explanations, and we evaluate the semantic closeness and paraphrase quality for $s_i, s_f$ with reference-free metrics. Similarly, we evaluate $e_b, s_n$ for the neutralizing subjective bias task using a test set from \textsc{e-WNC}. We report F1 scores for bias classification in $s_n$.

\subsection{Models}
We test two smaller student models fine-tuned on our datasets. We fine-tune LLaMA-7B \cite{touvron2023llama} and Alpaca-7B \cite{alpaca} models on our data converted to the Alpaca instruction format.
For formality style transfer, we fine-tune in both Formal$\rightarrow$Informal and Informal$\rightarrow$Formal directions separately ($\rightarrow$), as well as in both directions in a multi-task fashion ($\leftrightarrow$). For subjective bias transfer, we only fine-tune in one direction. In addition, we test the teacher models (ChatGPT-3.5=GPT-3.5 and ChatGPT-4=GPT-4) in few-show setting, which is an ambitious baseline: first, they were used for data generation which biases reference-based metrics, second, they are  prompted with improved instances of the data. Since the teacher models are closed models, we also test a representative open instruction-tuned model larger than the student, Vicuna-13B \cite{vicuna2023} (Vic in tables), in few-shot setting.
See detailed description of the models, hyperparameters, prompts, and additional experiments in Appendix \ref{appendix:models}.

\subsection{Automatic Evaluation} \label{subsec:metrics}
We use the following metrics:
\begin{itemize}
    \item BLEU \cite{papineni-etal-2002-bleu}: We measure the amount of exactly matched formal and informal attributes and evidences between the generated structured explanation and reference explanation in \textsc{e-GYAFC} and \textsc{e-WNC}.
    \item Mutual Implication Score (MIS) \cite{babakov-etal-2022-large} is a symmetric measure of text semantic similarity based on a RoBERTa \cite{liu2019roberta} model fine-tuned for natural language inference and paraphrase detection used in prior work \cite[e.g.,][]{patel2022lowresource}. 
    \item Style Accuracy: For \textsc{e-GYAFC}, we use Formality/Informality Score\footnote{\href{https://huggingface.co/s-nlp/roberta-base-formality-ranker}{huggingface.co/s-nlp/roberta-base-formality-ranker}}: RoBERTa \cite{liu2019roberta} fine-tuned to predict whether sentences are formal or informal using GYAFC and Online Formality Corpus (OFC) \cite{pavlick-tetreault-2016-empirical}. It achieves up to 0.98 ROC AUC. For \textsc{e-WNC}, we use Bias Score\footnote{\href{https://huggingface.co/Social-Media-Fairness/Classifier-Bias-SG}{huggingface.co/social-media-fairness/classifier-bias-sg}}: DistilBERT model \cite{sanh2020distilbert} fine-tuned for bias classification on the BABE media bias dataset annotated by experts \cite{spinde-etal-2021-neural-media}, on which it obtaines F1 score of up to 79.
\end{itemize}
 Given that the bias type detection task can be viewed as a classification task (only 3 labels are present), we report F1 scores for bias type classification in the explanation. We also report the average across all metrics.
 
\paragraph{Results} Table \ref{tab:instruct} shows model performance on the explainable formality style transfer task. While the Vicuna model does well in terms of style transfer (as evidenced by high MIS and Formality scores), it lacks in explanation quality (overall low BLEU scores). Student models perform better than the one-shot teacher model and competitively in 10-shot scenario judging by the Average score.
Table \ref{tab:bias} shows model performance on the explainable subjective bias style transfer task. Similarly, student models outperform the teacher model in one-shot setting. They also outperform few-shot models weaker than the teacher model (Vicuna and GPT-3.5).

As for style transfer performance without considering the explanations, we see a slight decrease in the Informal$\rightarrow$Formal (-0.93\% avg. MIS and Formality compared to one-shot teacher) and Bias$\rightarrow$Unbias (-2.13\% avg. MIS and Neutrality) tasks. This is expected as consistent with prior work such as e-SNLI \cite{NEURIPS2018_4c7a167b}.\footnote{``...while sacrificing a bit of performance, we get a better trust that when EXPLAINTHENPREDICT predicts a correct label, it does so for the right reasons.''} We see a sizeable increase in performance for Formal$\rightarrow$Informal (+12.60\%) direction. The informality scores are much better for the student models, perhaps due to the tendency to generate more formal speech both by teacher models and the other instruct models. 

\begin{table}[h]
\centering
\small
\begin{tabularx}{\columnwidth}{Xccccc}
\toprule
Model & F1 & \makecell{Attrs. \\ BLEU} & MIS & \makecell{Neutr.\\ score} & \textbf{Avg} \\
\midrule
Vic$_1$ & 17.49 & 2.88 & 74.61 & 68.33 & 48.61 \\
Vic$_5$ & 16.28 & 9.58 & 81.22 & 73.12 & 54.64\\
Vic$_{10}$ & 23.08 & 9.35 & 84.66 & 74.90 & 56.30 \\
GPT-3.5$_1$ & 34.83 & 8.15 & 83.18 & 75.07 & 55.47 \\
GPT-3.5$_5$ & 28.78 & 13.94 & 81.71 & 73.11 & 56.25 \\
GPT-3.5$_{10}$ & 37.83 & 17.25 & 82.92 & 73.18 & 57.78\\
GPT-4$_1$ & 36.87 & 12.33 & 82.38 & \textbf{75.72} & 56.81 \\
GPT-4$_5$ & 41.24 & 14.91 & 82.36 & 75.48 & 57.58 \\
GPT-4$_{10}$ & 39.82 & 17.07 & \textbf{83.48} & 74.87 & 58.47 \\
\midrule \midrule
Alpaca$_{\rightarrow}$ & 65.03 & 24.46 & 82.57 & 71.63 & 59.55 \\
LLaMA$_{\rightarrow}$ & \textbf{67.25} & \textbf{25.81} & \textbf{83.48} & 71.33 & \textbf{60.21} \\
\bottomrule
\end{tabularx}
\caption{Performance on Biased$\rightarrow$Unbiased explainable style transfer. GPT-3.5 = ChatGPT-3.5, GPT-4 = ChatGPT-4, Vic = Vicuna-13B. Number of shots in underscripts. Best score bolded.}
\label{tab:bias}
\end{table}

\subsection{Expert Human Evaluation: Preference Judgments}
We evaluate how well models' outputs align with expert preferences. For \texttt{e-GYAFC}, we sample 100 instances from the test set and the corresponding model outputs from Vic$_{1}$, 
the teacher model GPT-3.5 in one-shot setting (GPT-3.5$_{1}$), and the best student fine-tuned model. We ask for their preferences in terms of correctness and completeness of explanations, as well as in terms of paraphrase semantic preservation. We find that student model (Alpaca$_{\leftrightarrow}$) generations are preferred to teacher model GPT-3.5$_1$ and Vicuna$_1$ 53\% of the time by the expert linguist (GPT-3.5$_1$ is preferred 42\%), indicating the student model is more aligned with expert preferences. See a qualitative example in Table \ref{tab:modelgeneval}. We also evaluate how well do smaller fine-tuned models perform compared to the teacher model in a few-shot setting. We sample 40 outputs and ask for preferences between generations from Vicuna and GPT-3.5 in the 10-shot setting (GPT-3.5$_{10}$) as well as the student fine-tuned model. The fine-tuned model performs competitively with 49\% preference for GPT-3.5$_{10}$ and 43\% preference for the small student fine-tuned model.

Similarly, we hire an expert annotator and provide 50 test set instances from \textsc{e-WNC} along with model outputs from GPT-3.5$_{10}$ (weaker than the teacher model), 10-shot Vicuna, and the best student fine-tuned model (LLaMA$_{\rightarrow}$). We find that LLaMA$_{\rightarrow}$ is preferred at a rate of 56\%, followed by GPT-3.5$_{10}$ (24\%). See qualitative examples in Table \ref{tab:modelgenBias}. We also measure preferences to one-shot teacher model (GPT-4$_1$), finding comparable preference of 42\% for LLaMA$_{\rightarrow}$ and 40\% for GPT-4$_1$.

\begin{table}[h!]
\centering
\small
\begin{tabularx}{\columnwidth}{X}
\toprule
\textbf{Formal (Input)}\\
If it is feasible, allow love to prevail. \\
\midrule
\textbf{Vicuna$_1$} \\
\midrule
-\texttt{Formal Attributes:} passive voice ("it is feasible", "allow love to prevail"), use of verb "to be" ("is feasible", "prevail") \\
-\texttt{Informal Paraphrase:} It's okay to let love rule. \\
-\texttt{Informal Attributes:} active voice ("it's okay", "let love rule"), informal verb use ("rule") \\
\midrule
\textbf{GPT$_1$} \\
\midrule
-\texttt{Formal Attributes:} conditional clause ("if it is feasible"), elevated vocabulary ("allow love to prevail") \\
-\texttt{Informal Paraphrase:} If it's possible, just go with love. \\
-\texttt{Informal Attributes:} simple language ("if it's possible"), imperative form ("just go"), contraction ("it's") \\
\midrule
\textbf{Alpaca$_{\leftrightarrow}$} \\
\midrule
-\texttt{Formal Attributes:} precise language ("feasible"), use of conditional clause ("if it is feasible"), lexical sophistication ("prevail") \\
-\texttt{Informal Paraphrase:} let love win if it's possible. \\
-\texttt{Informal Attributes:} contraction ("it's"), casual tone ("let love win") \\
\bottomrule
\end{tabularx}
\caption{Comparison between generations from a one-shot instruction-tuned model (Vicuna, ChatGPT), and our best small student fine-tuned model for explainable formality style transfer.}
\label{tab:modelgeneval}
\end{table}

\begin{table}[h!]
\centering
\small
\begin{tabularx}{\columnwidth}{X}
\toprule
\textbf{Biased (Input)}\\
some cacti produce \textbf{beautiful} flowers, which like spines and branches arise from areoles. \\
-\texttt{Bias Attributes (Gold)}: Framing ("beautiful" adds subjectivity and implies a positive evaluation of the flowers)\\
\midrule
\textbf{Vicuna$_{10}$ \textcolor{red}{(Incorrect)}} \\
\midrule
-\texttt{Bias Attributes:} Framing (using "some" suggests that not all cacti produce beautiful flowers)\\
-\texttt{Neutralized Paraphrase:} Certain cacti produce beautiful flowers, which like spines and branches arise from areoles. \\
\midrule
\textbf{GPT-3.5$_{10}$ \textcolor{red}{(Incorrect)}} \\
\midrule
-\texttt{Bias Attributes:} Epistemological (using the word "beautiful" implies that the flowers produced by cacti are universally considered attractive) \\
-\texttt{Neutralized Paraphrase:} Some cacti produce flowers, which, like spines and branches, arise from areoles.\\
\midrule
\textbf{LLaMA$_{\rightarrow}$ \textcolor{green}{(Correct)}} \\
\midrule
-\texttt{Bias Attributes:} Framing ("beautiful" suggests a subjective evaluation of the flowers) \\
-\texttt{Neutralized Paraphrase:} Some cacti produce flowers, which like spines and branches arise from areoles. \\
\bottomrule
\end{tabularx}
\caption{Comparison between generations from a few-shot instruction-tuned models (10-shot Vicuna, ChatGPT-3.5), and our best small student fine-tuned model for explainable bias style transfer.}
\label{tab:modelgenBias}
\end{table}

\section{Extrinsic Evaluation of Formality Style Transfer Explanations} \label{sec:apps}

We use Authorship Verification \cite{authorattributionstyle, coulthard2004author, sylometrytech} as an extrinsic task utilizing PAN 2022 \cite{bevendorff2022overview} data. This is a binary classification  task of deciding if two texts belong to the same author or not. The two input texts are two raw documents (e.g., paragraphs from blog posts), one of which is written by author A, and another one which is either written by the same author A or by a different author B. We then derive a representation (features to be used for classification) of these input texts using our explainable style transfer model. We run Alpaca$_{ \text{IF} \rightarrow \text{ F } }$ on each text (at most 15 sentences per author are considered) and extract explanations containing informality attributes and evidence (see Table \ref{tab:attrs}). In a preliminary evaluation of the usefulness of these features, we compute the similarity between authors by measuring the percentage of overlapping attributes. Note that the evidence fragments corresponding to the attributes are not used in this preliminary experiment. We then use the percentage of overlapping attributes as a classification score, for example, if author A uses colloquialism and textese, and author B uses colloquialism and abbreviation, their similarity score is the number of common attributes divided by the number of unique attributes between the authors, or $\frac{1}{3}$ (0.33). Here, only 1 attribute is common (colloquialism) whereas the total identified attributes is 3 (colloquialism, abbreviation, textese). Following the computation described above, we take the similarity score as a confidence for a binary prediction task. If the similarity is high, there is a high chance the authors are the same (prediction = 1), and if not they are likely not the same (prediction = 0). The underlying assumption is that it is more likely that the same author would use some of the informality features they used previously (not every sentence in PAN is completely informal, but informality is a very broad category, so when authors do use some informal attributes they can provide a signal for authorship). We compute ROC AUC between the confidence scores (author similarity score) and ground truth predictions from the PAN dataset (1 if authors are the same and 0 if not). We compare explanations from Vicuna$_{10}$, GPT-3.5$_{10}$ and Alpaca$_{ \text{IF} \rightarrow \text{ F } }$ by their predictive signal for this task. Explanations by Alpaca$_{ \text{IF} \rightarrow \text{ F } }$ achieve an AUC of 56.4, whereas explanations from the Vicuna and GPT-3.5 models achieve a score of 50.0 and 47.0 respectively. This indicates a potential application of the explanations generated by the student model fine-tuned on our dataset (\textsc{e-GYAFC}) to be used as interpretable authorship features that can be explored in future work.

\begin{table}[h!]
\centering
\small 
\begin{tabularx}{\columnwidth}{l|X}
\hline
\textbf{Attribute} & \textbf{Evidence} \\ 
\hline
Colloquialism & ``assumed they all started off low!?'', ``typing it out'' \\
\hline
Textese & ``xx'' \\
\hline
Informal Vocabulary & ``give you a call'', ``arrange something'' \\
\hline
Informal Tone & ``hoping to borrow a couple of charging leads'' \\
\hline
\end{tabularx}
\caption{Informality features for authorship identification: on the left, informality attributes identified by our model, on the right, textual evidence provided by it.}
\label{tab:attrs}
\end{table}

\section{Conclusion}
We propose a framework to augment two style transfer datasets with semi-structured textual explanations. To improve quality of model distillation and incorporate expert feedback, we propose \texttt{ICLEF} (In-Context Learning from Expert Feedback), a novel human-AI collaboration framework leveraging both in-context learning and self-critique abilities of LLMs. We evaluate smaller student models fine-tuned on the resulting datasets compared to large teacher models and conduct expert human evaluation. We also extrinsically evaluate the explanations for the formality style transfer on the downstream task of
authorship attribution.

\section{Limitations}
The GYAFC dataset does not contain all types of informal and formal language, namely, they mostly focus on interpersonal relationships (the subset used for this paper) and entertainment. Future work could consider extending our approach to other style transfer datasets, including ones more encompassing of formality.

While our methods are intended to produce faithful explanations, there can still be instances when a model does not rely on the attributes in order to complete the paraphrase. We also observed that hallucinations can still be present in our fine-tuned models' explanations and hope that future work will try to address these issues. We also note that our approach does not replace expert annotation as it heavily relies on LLMs that may still hallucinate. It is only meant to be applicable in scenarios where expert feedback is expensive and/or difficult to gather.

One limitation of our method is that we used a relatively small number of experts to conduct our study. However, we believe that this setting mirrors real-life conditions where experts are usually scarcely available. We hope our approach provides a more general framework for incorporating expert feedback that can be adjusted to experts' needs (e.g., a forensic linguist may require a different style transfer explanation than a literary critic). 

Fine-tuning and running inference on large models requires expensive computational resources. However, we hope that our study presents a convincing argument that fine-tuning a smaller model once may be more efficient and accurate than running a large general-purpose model with elaborate long-context prompts.

\section{Ethics Statement}
The GYAFC corpus was created using the Yahoo Answers corpus: L6 - Yahoo! Answers Comprehensive Questions and Answers version 1.0. This Yahoo Answers corpus can be requested free of charge for research purposes. Access to our GYAFC dataset will require users to first gain access to this Yahoo Answers corpus. Authors obtained permission to access the dataset.

Our datasets do not include any protected data to the best of our knowledge.
All annotators are fairly compensated for their work in accordance with their asking rate (typically over 20 USD per hour).

Our bias style transfer model is only intended for use in a human-in-the-loop fashion and not by itself to adjudicate bias in text. We hope that the explanation generation capacity of  our model will improve upon existing bias classifiers that typically do not provide textual explanations.
In Appendix \ref{app:aigentxt}, we show how style transfer can be used to evade AI-text detectors. Similarly to \citet{krishna2023paraphrasing}, we reiterate that this is not to provide a way to attack such systems, but to bring awareness to the community that current detectors are easy to evade. Moreover, we bring to attention the issue of detecting text on which style transfer paraphrase has been applied. We hope that future work develops systems capable of defending against such attacks, perhaps utilizing explanations generated by our system.

GYAFC and WNC may potentially contain offensive data as they are crowdowdsourced, however, on samples that we saw we did not find alarming ethical issues.

\section*{Acknowledgements}
We thank the annotators for their work and providing detailed feedback. We also would like to thank the reviewers for productive and engaging discussions. This research is supported in part by the Office of the Director of National Intelligence (ODNI), Intelligence Advanced Research Projects Activity (IARPA), via the HIATUS Program contract \#2022-22072200005. The views and conclusions contained herein are those of the authors and should not be interpreted as necessarily representing the official policies, either expressed or implied, of ODNI, IARPA, or the U.S. Government. The U.S. Government is authorized to reproduce and distribute reprints for governmental purposes notwithstanding any copyright annotation therein.

\bibliography{anthology,custom}
\clearpage
\appendix

\section{Human evaluation details: \textsc{e-GYAFC}} \label{app:human}

We hire three annotators for preference evaluation: A1 (bachelors degree in linguistics), A2 (bachelors degree in linguistics and a masters degree in education), A3 (bachelor
and master degrees in linguistics) on Upwork. Due to high prevalence of the positive class, there is a high chance of random agreement, hence we provide a more granular look into the expert annotations than the inter-rater agreement in Table \ref{tab:humaneval_egyafc}. Pairwise accuracy between annotator responses for all categories of \textsc{e-GYAFC} evaluation, and found that it averages at 81\% across all categories. Annotator A2 expressed concerns that the paraphrases may sound unnatural due to excessive formality (we believe it is due to the context in which the informal expression would be uttered) and that explanations sometimes miss punctuation errors (which, while important, is not critical for model-generated explanations).

\begin{table}[ht]
\begin{adjustbox}{width=0.95\columnwidth,center}
\centering
\begin{tabular}{lccccc}
\toprule
 & \makecell{$e_i$ \\ pref.} & \makecell{$e_i$ \\ accept.} & \makecell{$s_f$ \\ pref.} & \makecell{$s_f$ \\ accept.} & \makecell{$e_f$ \\ accept.} \\
\midrule
A1 & 91\% & 95\% & 64\% & 64\% & 98\%\\
A2 & 87\% & 84\% & 76\% & 75\% & 96\% \\
A3 & 91\% & 83\% & 91\% & 93\% & 100\% \\
\bottomrule
\end{tabular}
\end{adjustbox}
\caption{Expert evaluation of \textsc{e-GYAFC} dataset quality. We report percentage of time each item was preferred, as well as acceptability judgements.}
\label{tab:humaneval_egyafc}
\end{table}

\section{Human evaluation details: \textsc{e-WNC}} \label{app:humanBias}

For \textsc{e-WNC}, we hire two expert annotators A1 and A2. 
A1 is a professional with over ten years experience in translation and interpretation, data annotation, linguistics and publishing. A2 is a PhD in linguistics with background in psycholinguistics and neurolinguistics and experience in writing, proofreading and editing academic texts

\begin{table}[ht]
\begin{adjustbox}{width=0.95\columnwidth,center}
\centering
\small
\begin{tabular}{lcccc}
\toprule
 & \makecell{$e_b$ \\ pref.} & \makecell{$e_b$ \\ accept.} & \makecell{$s_n$ \\ pref.} & \makecell{$s_n$ \\ accept.} \\
\midrule
A1 & 82\% & 84\% & 84\% & 86\% \\
A2 & 74\% & 62\% & 70\% & 62\%  \\
\bottomrule
\end{tabular}
\end{adjustbox}
\caption{Expert evaluation of \textsc{e-WNC} dataset quality. We report percentage of time each item was preferred, as well as acceptability judgements.}
\label{tab:humaneval_ewnc}
\end{table}

\section{How does ICLEF performance change depending on the amount of feedback provided?} \label{app:iclefNumFeed}
\begin{figure}[h!]
    \centering
    \includegraphics[width=\columnwidth]{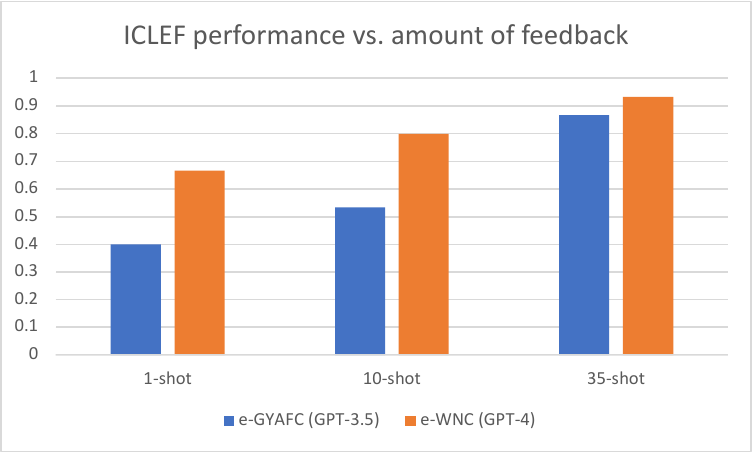}
    \caption{ICLEF performance increases with amount of feedback, reaching satisfactory accuracy at around 35 shots.}
    \label{fig:iclefPerf}
\end{figure}

We perform a small-scale study to explore how performance of the self-critique component of ICLEF changes depending on the number of in-context examples provided. We sample 15 instances of synthetic explanations (before applying ICLEF) and evaluate generated critiques for correctness when providing 1, 10, or 35 instances of expert feedback few-shot. We evaluate the critiques for correctness, i.e. they have to both identify the incorrect attribute and not introduce any incorrect attributes. Figure \ref{fig:iclefPerf} shows how by increasing the amount of feedback to 35-shot, the correctness is raised to $\approx$ 87\% for \textsc{e-GYAFC} and $\approx$93\% for \textsc{e-WNC}, which can be deemed satisfactory.

\section{Additional estimations for trustworthiness of explanations}
We conduct expert evaluation of our data in Section \ref{section:dataQuality}. To further confirm trustworthiness of explanations, we check that every evidence fragment actually occurs in the text. For every data instance, we compute what percentage of evidences was present in the sentence. We report average across the respective datasets and explanations in Table \ref{tab:trustworthiness} summarizes the results, where it is evident that the explanations have high trustworthiness. We note that some explanations discuss absence of certain words in which case they should not be present in the sentence in the first place. Moreover, our structured explanation format would make it very easy for the end-user to verify the correctness of the explanation and alert them of potentially incorrect paraphrase.

\begin{table}[h]
\small
\centering
\begin{tabular}{@{}lccc@{}}
\toprule
& \multicolumn{2}{c}{\textbf{e-GYAFC}} & \textbf{e-WNC} \\
\cmidrule(lr){2-3} \cmidrule(lr){4-4}
& \textbf{e$_i$} & \textbf{e$_f$} & \textbf{e$_u$} \\
\midrule
Trustworthiness & 95.0\% & 92.3\% & 98.9\% \\
\bottomrule
\end{tabular}
\caption{Explanation Trustworthiness Rates Across Datasets: percentage of evidences actually present in the sentence.}
\label{tab:trustworthiness}
\end{table}

\section{Prompts}\label{app:prompts}

\begin{table}[h]
\centering
\small
\begin{tabular}{|p{0.93\columnwidth}|}
\hline
\textit{You are an extremely attentive and critical annotator with background in stylometry and linguistics.} You will be provided with an informal sentence. You will also be provided with an explanation of its informality attributes. 
Decide whether the explanation is incorrect, and if so, describe what attributes were listed incorrectly. 
\\
EXAMPLES:
<…>\\
\hline
\textit{You are an extremely attentive and critical annotator with background in ethics, journalism, critical thinking and bias identification.} You will be provided with a possibly biased sentence. You will also be provided with an explanation of its bias attributes. <...> If the explanation is incorrect, reply with a correction. Focus on three main types of bias in a sentence: <…> \\
EXAMPLES:
<…>\\
\hline
\end{tabular}
\caption{Prompts for LLM-critic models. Top is used for formality style transfer, bottom is used for subjective bias style transfer.}
\label{tab:prompts}
\end{table}

\paragraph{ChatGPT Explanation Generation Prompt}
We provide an instruction as a system prompt (``You are an expert forensic linguist...'') and 6 examples of the task in the OpenAI ChatML format.\footnote{\href{https://github.com/openai/openai-python/blob/main/chatml.md}{github.com/openai/openai-python/blob/main/chatml.md}}

\texttt{Instruction: You are an expert forensic linguist. Your task is to identify infromal attributes in a setnece, modify them to create a formal sentence, and then output the attributes of the generated formal sentence. Use the following format: attribute (excertp from text in quotation marks). Make sure to provide a complete list of informal and formal atttributes. Focus on what has changed between formal and informal sentences. Informal writing tends to be more casual, personal, and conversational than formal writing. Here are some common features of informal writing: Contractions: Informal writing often uses contractions, such as "I'm," "can't," "won't," and "they've," which are generally avoided in formal writing. ...}

\texttt{Examples: Informal greetings and sign-offs: Informal writing often uses casual greetings, such as "Hi" or "Hey," and sign-offs like "Cheers" or "Take care." Informal: if ur under 18 u have a BIG PROBLEM. Informal Candidates: {'18', 'BIG', 'PROBLEM.', 'if', 'u', 'ur} Attributes of Informal Style: textese ("ur", "u"), capitalization ("BIG PROBLEM"), colloquialism ("BIG PROBLEM")...
}

\paragraph{Prompt for ChatGPT ICLEF generation} 
\texttt{You are an extremely attentive and critical annotator with background in stylometry and linguistics. You will be provided with an informal sentence. You will also be provided with an explanation of its informality attributes. 
Decide whether the explanation is incorrect, and if so, describe what attributes were listed incorrectly. }

\texttt{EXAMPLES: Informal Sentence: Look,  If you really like this person,  just tell her.
Informal Attributes: colloquialism ("just tell her"), contraction ("If you"), simple sentence structure. Attributes Listed Incorrectly: contraction ("If you" is not a contraction)...}

\section{Model details, additional models and experiments} \label{appendix:models}
Below we provide some additional experiments and models used for them as well as experimentation details.

\paragraph{Instruction-tuned Models} All instruction-tuned models are provided with the same one-shot prompt (modulo special token requirements) and generation parameters.
\begin{itemize}
    
    \item MPT-7B-Instruct: built by finetuning MPT-7B \cite{MosaicML2023Introducing} on a dataset derived from the Databricks Dolly-15k \cite{dolly} and the Anthropic Helpful and Harmless \cite{bai2022training} datasets. 
    \item Alpaca-7B \cite{alpaca} a model fine-tuned from the LLaMA-7B model on 52K instruction-following demonstrations generated with the Self-Instruct framework \cite{wang-etal-2023-self-instruct}.
    \item Vicuna-13B \cite{vicuna2023}: an open-source chatbot trained by fine-tuning LLaMA on user-shared conversations collected from ShareGPT\footnote{\href{https://sharegpt.com/}{sharegpt.com}}. It places first in the Huggingface Open LLM Leaderboard \cite{open-llm-leaderboard} based on human and GPT-4 evaluation as of writing this paper.
    \item Falcon-40B \cite{falcon40b} causal decoder-only model trained on 1,000B tokens of RefinedWeb \cite{refinedweb} enhanced with curated corpora.
    \item Tülu-65B \cite{wang2023far} a 65B LLaMa model finetuned on a mixture of instruction datasets (FLAN V2 \cite{longpre2023flan}, CoT \cite{wei2022chain}, Dolly \cite{dolly}, Open Assistant 1 \cite{köpf2023openassistant}, GPT4-Alpaca \cite{peng2023instruction}, Code-Alpaca \cite{codealpaca}, and ShareGPT).
\end{itemize}

\paragraph{ChatGPT} For ChatGPT-3.5 we use \texttt{gpt-3.5-turbo-1106}. For ChatGPT-4 we use \texttt{gpt-4}.

\paragraph{Fine-tuned models}
We fine-tune below models on \textsc{e-GYAFC}. $\rightarrow$ indicates fine-tuning two models in each direction, and $\leftrightarrow$ indicates fine-tuning on combined data in both directions.
\begin{itemize}
    \item FLAN-T5-XL$_{\leftrightarrow}$ \cite{chung2022scaling} approximately 3B parameter instruction-tuned model based on the T5 architecture \cite{t5txt}.
    \item LLaMA-7B$_{\rightarrow}$ \cite{touvron2023llama} model by Meta trained on 1 trillion tokens.
    \item Alpaca-7B$_{ \rightarrow, \leftrightarrow  }$ \cite{alpaca} a model fine-tuned from the LLaMA-7B model on 52K instruction-following demonstrations. In addition, we test Alpaca-7B$_{ noexpl }$ as the model fine-tuned for Formal to Informal style transfer with no explanations provided in the fine-tuning data or in the output.
\end{itemize}

\paragraph{Fine-tuning hyperparameters} We fine-tune all models using the script provided in the Stanford Alpaca repository.\footnote{\href{https://github.com/tatsu-lab/stanford_alpaca}{Alpaca GitHub}} We use exact same hyperparameters, except for batch size which we adjust to 1 due to memory constraints.  We fine-tune our models on 4 A100 NVIDIA 40GB GPUs. We train our models for 3 epochs with learning rate  2e-05 with cosine rate scheduler and warmup ratio of 0.03. We did not perform hyperparameter search. We report results form single runs with random seeds preserved due to computational constraints.

\paragraph{Inference parameters}
We use the same hyperparameters for generation across all models, that is temperature=0.7, top p=0.9, max new tokens = 256. We use the huggingface library.

One-shot instruction is provided below:
\texttt{Identify informal attributes in a given sentence, modify them to create a formal sentence, and then output the attributes of the generated formal sentence.}

\texttt{For example:}

\texttt{Informal: how can you tell if a girl likes you or not?}

\texttt{Informal Attributes: direct question form ("how can you tell"), informal language ("girl", "likes you") Formal: What are some indications that a woman may be interested in you? Formal Attributes: indirect question form ("what are some indications"), lexical sophistication ("woman", "interested in you")}

\texttt{For the following sentence, identify informal attributes in a given sentence, modify them to create a formal sentence, and then output the attributes of the generated formal sentence.}

For Tulu, we add the <asistant> and <user> tokens as advised by model developers.

\paragraph{Packages} We used transformers \cite{wolf2020huggingfaces} for language model inference and NLTK package \cite{bird-loper-2004-nltk} for sentence tokenization.

\paragraph{Additional experiments with instruct and fine-tuned models}
We additionally perform experiments with MPT-7B, Falcon-40B, Tulu, and non-fine-tuned Alpaca. We also fine-tune an Alpaca model with no explanations. See Table \ref{tab:additionalExps}. The model fine-tuned without explanations (Alpaca$_{ noexpl }$) achieves comparable performance, indicating that generating explanations does not significantly hurt performance on the standard style transfer task.

\begin{table*}[h]
\begin{adjustbox}{width=\textwidth,center}
\centering
\small
\begin{tabular}{lccccccccccc}
\toprule
 & & \multicolumn{4}{c}{Formal $\rightarrow$ Informal} & \multicolumn{4}{c}{Informal $\rightarrow$ Formal} & \\
\cmidrule(lr){3-6} \cmidrule(lr){7-10}
Model & Size & \makecell{Form.Attrs. \\ BLEU} & \makecell{MIS} & \makecell{Informality} & \makecell{Inform.Attrs. \\ BLEU} & \makecell{Inform.Attrs. \\ BLEU} & \makecell{MIS} & \makecell{Formality} & \makecell{Form.Attrs. \\ BLEU} & \textbf{Average} \\
\midrule
\makecell[l]{MPT} & 7B & 24.59 & 51.84 & 9.82 & 2.10 & 23.26 & 46.26 & 58.40 & 0.86 & 27.14\\
\makecell[l]{Alpaca} & 7B &  17.07 & 80.74 & 32.53 & 6.51 & 23.67 & 73.69 & 86.82 & 8.27 & 41.16 \\
\makecell[l]{Falcon} & 40B &  8.38 & 28.12 & 13.43 & 1.23 & 20.80 & 38.01 & 62.69 & 7.13 & 22.47\\
\makecell[l]{Tülu} & 65B & 24.90 & 19.60 & 7.12 & 0.02 & 27.76 & 26.69 & 27.34 & 0.28 & 16.71 \\
\midrule \midrule
\makecell[l]{FLAN-T5$_{\rightarrow}$} & 3B & 0.00 & 8.54 & 0.01 & 0.00 & 0.00 & 9.82 & 0.91 & 0.00 & 2.41\\
\makecell[l]{Alpaca$_{\rightarrow}$} & 7B & 39.98 & 84.70 & 61.99 & 19.22 & 40.56 & 81.69 & 97.96 & 24.71 & 56.35\\
\makecell[l]{Alpaca$_{ noexpl }$} & 7B & - & 85.34 & 54.75 & - & - & 83.20 & 91.10 & - & - \\
\bottomrule
\end{tabular}
\end{adjustbox}
\caption{Performance of additional instruction-tuned and fine-tuned models on the explainable formality style transfer task.}
\label{tab:additionalExps}
\end{table*}

\paragraph{Fluency of style transfer} \label{sec:fluency}
Fluency could be used as an additional metric to measure the quality of the style transfer. However, fluency measures negatively correlate with informality, even though the semantic content stays the same. Because of that, we preferred to use semantic similarity of the output to the input (MIS) as the primary measure across all tasks. Below is the result of an experiment we conducted with perplexity (lower is better), following \citet{suzgun-etal-2022-prompt} exactly to compute the fluency metric.

\begin{table}[h!]
\centering
\small 
\begin{tabular}{lccc}
\toprule
Model & PPL Inf. $\rightarrow$ & PPL For. $\rightarrow$ & PPL Bias $\rightarrow$ \\
      & For.                  & Inf.                   & Unbiased \\
\midrule
gold & 33.32 & 68.96 & 31.26 \\
gpt1 & 31.24 & 21.39 & 28.21 \\
gpt5 & 34.83 & 21.10 & 27.59 \\
gpt10 & 32.55 & 23.82 & 27.62 \\
best model & 30.21 & 38.61 & 31.97 \\
\bottomrule
\end{tabular}
\caption{Fluency evaluation results. GPT refers to the teacher model, best model refers to Alpaca for formality transfer and LLaMA for bias transfer.}
\label{tab:fluency}
\end{table}

As can be seen from Table \ref{tab:fluency}, for formal and unbiased paraphrase the perplexity is comparable with the teacher in 1-shot and few-shot settings as well as the gold reference, whereas for informality the gold data has the worst perplexity, making this metric unfit when we want to transfer from formal to informal style. We want to add that we performed a human evaluation in Section 4.3 that includes the evaluation of the transferred sentences.

\paragraph{Performance of PromptRerank Baseline} \label{sec:promptRerank}
We considered including a PromptRerank baseline \cite{suzgun-etal-2022-prompt} as it is one of the previous SOTA approaches. On our observation the models did not perform competitively possibly because the models described in the paper are smaller than the contemporary LLMs. Future work may explore adjustments on this baseline to adapt it to the new models. Performance is depicted in Table \ref{tab:PRperformance}.

\begin{table*}[h!]
\centering
\small
\begin{tabular}{lcccc}
\toprule
Model & MIS (Formal to & Informality & MIS (Informal to & Formality \\
      & Informal)      &            & Formal)          &           \\
\midrule
Alpaca$_{\leftrightarrow}$ & 81.76 & 66.71 & 79.43 & 98.57 \\
promptRerank & 63.33 & 33.11 & 62.70 & 20.00 \\
\bottomrule
\end{tabular}
\caption{Model performance on style transfer tasks.}
\label{tab:PRperformance}
\end{table*}

\paragraph{BLEU and BERTScore between transferred text and gold references} \label{sec:BleuBert}

We opted for reference-free metrics since they would be less biased toward producing output that most closely matches the reference. However, we now ran the experiment with BLEU and BERTScore between transferred text (output) and gold reference from our e-GYAFC and e-WNC datasets. BERTScore is based on contextualized-embeddings and usually preferred to using BLEU-4 (which is based on 4-gram overlap) for evaluation of text generation and paraphrases. Results are shown below in Table \ref{tab:bleuRef}.

\begin{table*}[h!]
\centering
\small
\begin{tabular}{lcccccc}
\toprule
Model & \multicolumn{2}{c}{Formal $\rightarrow$ Informal} & \multicolumn{2}{c}{Informal $\rightarrow$ Formal} & \multicolumn{2}{c}{Bias $\rightarrow$ Unbiased} \\
      & BLEU & BERTScore & BLEU & BERTScore & BLEU & BERTScore \\
\midrule
gpt1        & 15.01 & 78.40 & 28.10 & 84.29 & 60.34 & 91.30 \\
gpt5        & 15.36 & 78.74 & 28.71 & 84.48 & 64.85 & 91.78 \\
gpt10       & 14.23 & 78.14 & 29.13 & 84.62 & 65.98 & 92.13 \\
best model & 15.33 & 77.57 & 20.75 & 80.78 & 70.49 & 93.23 \\
\bottomrule
\end{tabular}
\caption{Performance comparison of models on BLEU and BERTScore metrics with the gold reference. GPT refers to the teacher model, best model refers to Alpaca$_{\leftrightarrow}$ for formality transfer and LLaMA$_{\rightarrow}$ for bias to unbiased transfer.}
\label{tab:bleuRef}
\end{table*}

We see that using BERTScore, the student model is comparable to the teacher model and both are semantically close to the reference. BLUE scores are lower due to more stricter n-gram overlap requirements but that is across all models and scores are again comparable between smaller student models and the larger teacher models.

\section{Discussion on applications of models fine-tuned for explainable style transfer} \label{app:aigentxt}
\paragraph{Explainable formal $\rightarrow$ informal style transfer is an interpretable adversarial attack on AI-generated text detection methods, including retrieval} \citet{krishna2023paraphrasing} established that paraphrasing easily evades the detection of AI-generated text, and proposed a retrieval-based defense. However, we hypothesize that retrieval-based metrics will degrade as similarity between generations becomes more ambiguous, as is the case for formality style transfer. For example, an adversarial agent might generate a post containing misinformation in typical ``formal" language generated by a language model like ChatGPT. This text is relatively detectable by current classifiers and 100\% detectable by retrieval-based methods. However, the agent might apply a style transfer model to lower the formality of the message. Alarmingly, not only this accomplishes the goal of spreading the AI-generated message more effectively as the result looks more like user-generated text, but, as we show, it also decreases the chances of being detected as AI-generated by current methods.

We test this in the following setting: we use an online dataset of political tweets \footnote{\href{https://www.kaggle.com/datasets/crowdflower/political-social-media-posts}{kaggle.com}}, and sample 30 of them. We ask ChatGPT to generate a political commentary post on the topic of the tweet (GPT-F), as well as an informal paraphrase of the said post (GPT-Inf). We manually annotate the resulting summaries and select those that look like they could be legitimate political messages posted on social media and have valid paraphrases. We then use our Alpaca$_{ \text{F} \rightarrow \text{IF} }$ model to generate an informal paraphrase of the GPT-F posts sentence-by-sentence. We also verify that these paraphrases are semantically valid and close to the original GPT-Formal post and select 24 high-quality generations. We choose a relatively small sample since we want to verify the paraphrase was still close to the original sentence manually to ensure semantic control for the experiment.

We report detection scores\footnote{Since we do not have the human baseline text, we do not report the performance at 1\% FPR, but for our study it is sufficient to show the scores decrease in absolute terms.} from 4 methods surveyed by \citet{krishna2023paraphrasing}: GPTZero \cite{Tian}, OpenAI classifier \cite{OpenAI}, DetectGPT \cite{mitchell2023detectgpt}, and their proposed retrieval methods based on BM25 \cite{INR-019} or P-SP \cite{wieting-etal-2022-paraphrastic} retrievers. As can be seen in Table \ref{tab:text_detection}, the formal-to-informal transfer model significantly decreases detection scores of all AI-generated text detection methods, including the retrieval-based one (despite the fact that the retrieval corpus is significantly smaller than it would be in real-world). Interestingly, for the BM25 retrieval method, the ChatGPT paraphrases are slightly harder to detect than Alpaca$_{ \text{F} \rightarrow \text{ IF } }$, whereas it is easier for all other methods. Since we used ChatGPT to generate the original posts, we could not use the watermarking methods \cite{kirchenbauer2023watermark}, but this can be explored in future work.

This result highlights the need to investigate new methods of detecting style transferred AI-generated text. As formality style transfer remains an effective attack, informality features produced by our model could help improve such classifiers. We leave this investigation for future work.

\begin{table}[H]
\begin{adjustbox}{width=\columnwidth,center}
\centering
\begin{tabular}{lccccc}
\toprule
Models & GPTZero & OpenAI & GPTDetect  & BM25 & P-SP \\
\midrule
GPT-F & 85.92 & 70.64 & 104.88 & 100 & 100 \\
GPT-Inf & 69.58 & 54.24 & 65.42 & \textbf{48.15} & 74.99 \\
$\text{F}\rightarrow\text{IF} $ & \textbf{6.11} & \textbf{44.86} & \textbf{54.92} & 58.68 & \textbf{74.08} \\
\bottomrule
\end{tabular}
\end{adjustbox}
\caption{Performance of various AI-generated text detectors on informal paraphrases from our model. Even retrieval methods perform poorly in this setting.}
\label{tab:text_detection}
\end{table}

\section{\textsc{e-EGYAFC} Statistics}

\begin{figure}[h]
    \centering
    \includegraphics[width=\columnwidth]{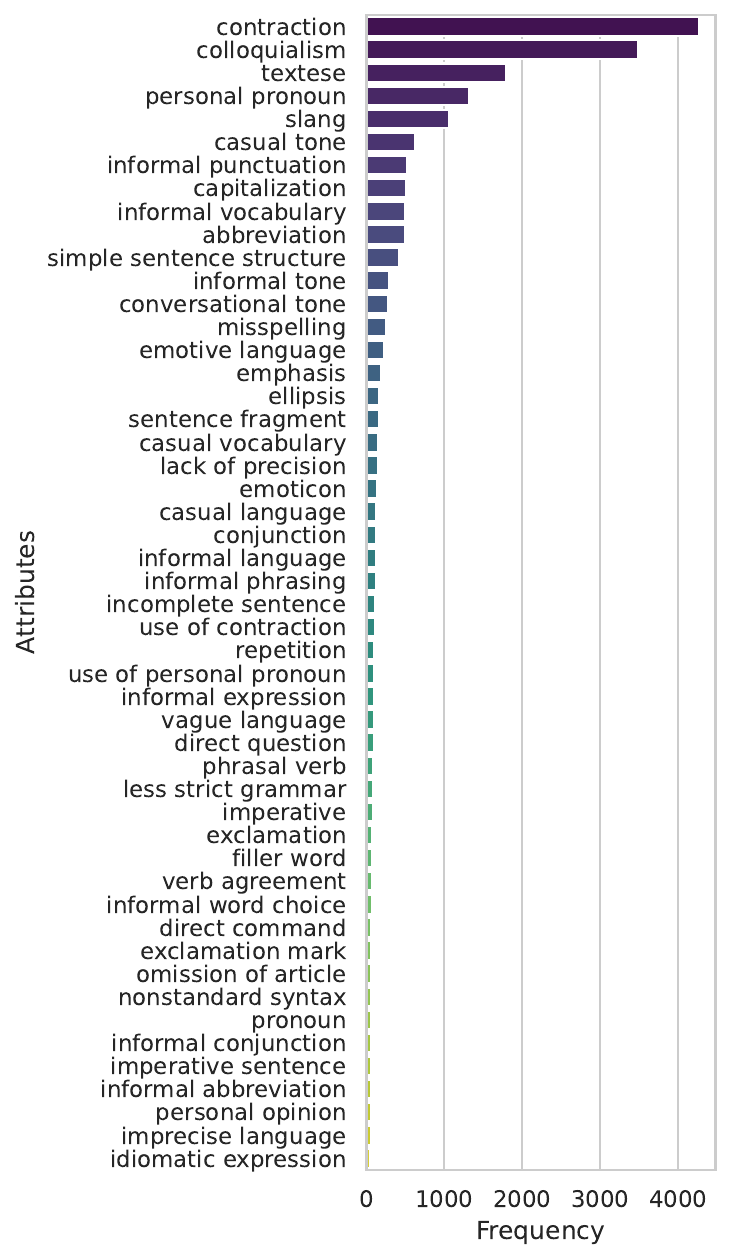}
    \caption{Distribution of 50 most frequent informal attributes in the \textsc{e-GYAFC} dataset.}
    \label{fig:informalDistr}
\end{figure}

\begin{figure}[h]
    \centering
    \includegraphics[width=\columnwidth]{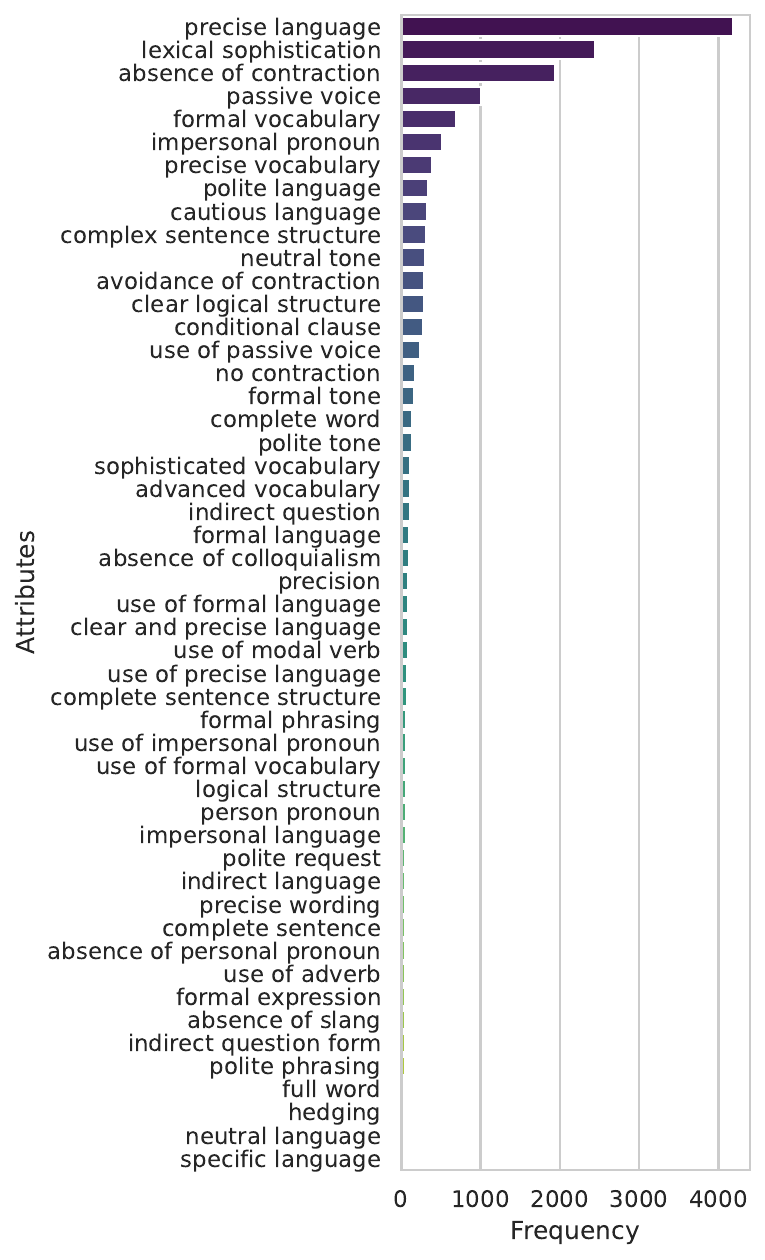}
    \caption{Distribution of 50 most frequent formal attributes in the \textsc{e-GYAFC} dataset.}
    \label{fig:formalDistr}
\end{figure}

We provide the distribution of 50 most frequent informal and formal attributes in \textsc{e-GYAFC} in Figures \ref{fig:informalDistr}, \ref{fig:formalDistr}.

\section{\textsc{e-WNC} Statistics} \label{app:biasStats}

We provide a proportion of classes in \textsc{e-WNC} in Table \ref{tab:ewncStats}.

\begin{table}[h]
    \centering
    \begin{tabular}{@{}lc@{}}
        \toprule
        Category        &  (\%) \\ 
        \midrule
        Demographic     & 3.70       \\
        Epistemological & 22.87      \\
        Framing         & 67.53      \\
        No Bias         & 5.90       \\
        \bottomrule
    \end{tabular}
    \caption{Proportion of classes in \textsc{e-WNC}}
    \label{tab:ewncStats}
\end{table}

\section{Annotation protocols} \label{app:protocols}
\begin{figure*}[h]
    \centering
    \includegraphics[scale=0.4]{ 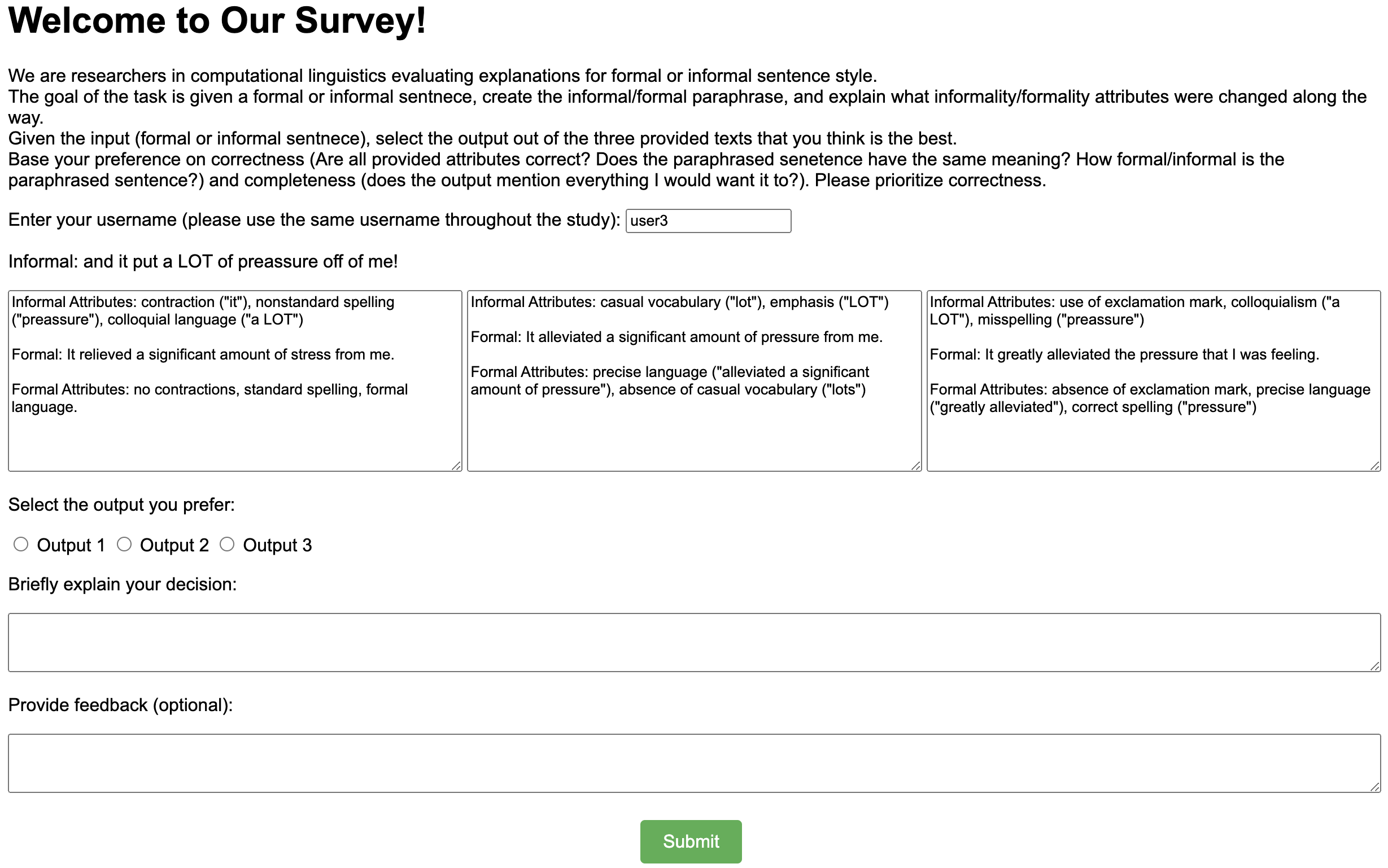}
    \caption{Annotation for model preferences.}
    \label{fig:modelgen-annot}
\end{figure*}

The screenshots for explanation interfaces are provided below in Figures \ref{fig:iclef-annot}, \ref{fig:egyafc-annot}, \ref{fig:modelgen-annot}. Similar annotation interfaces were used for the bias task.

\begin{figure*}[h]
    \centering
    \includegraphics[scale=0.6]{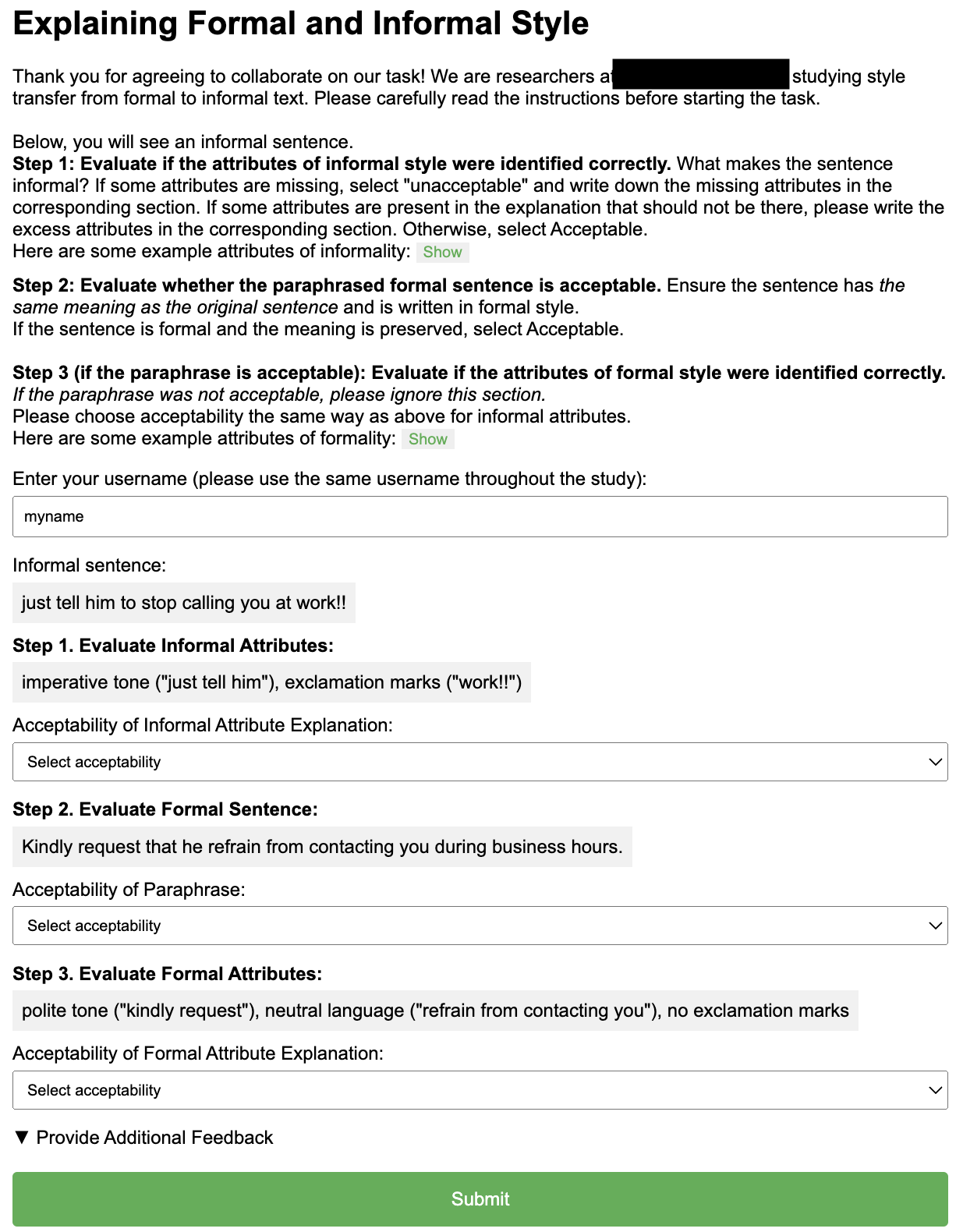}
    \caption{Annotation to gather feedback for ICLEF.}
    \label{fig:iclef-annot}
\end{figure*}

\begin{figure*}[h]
    \centering
    \includegraphics[scale=0.6]{ 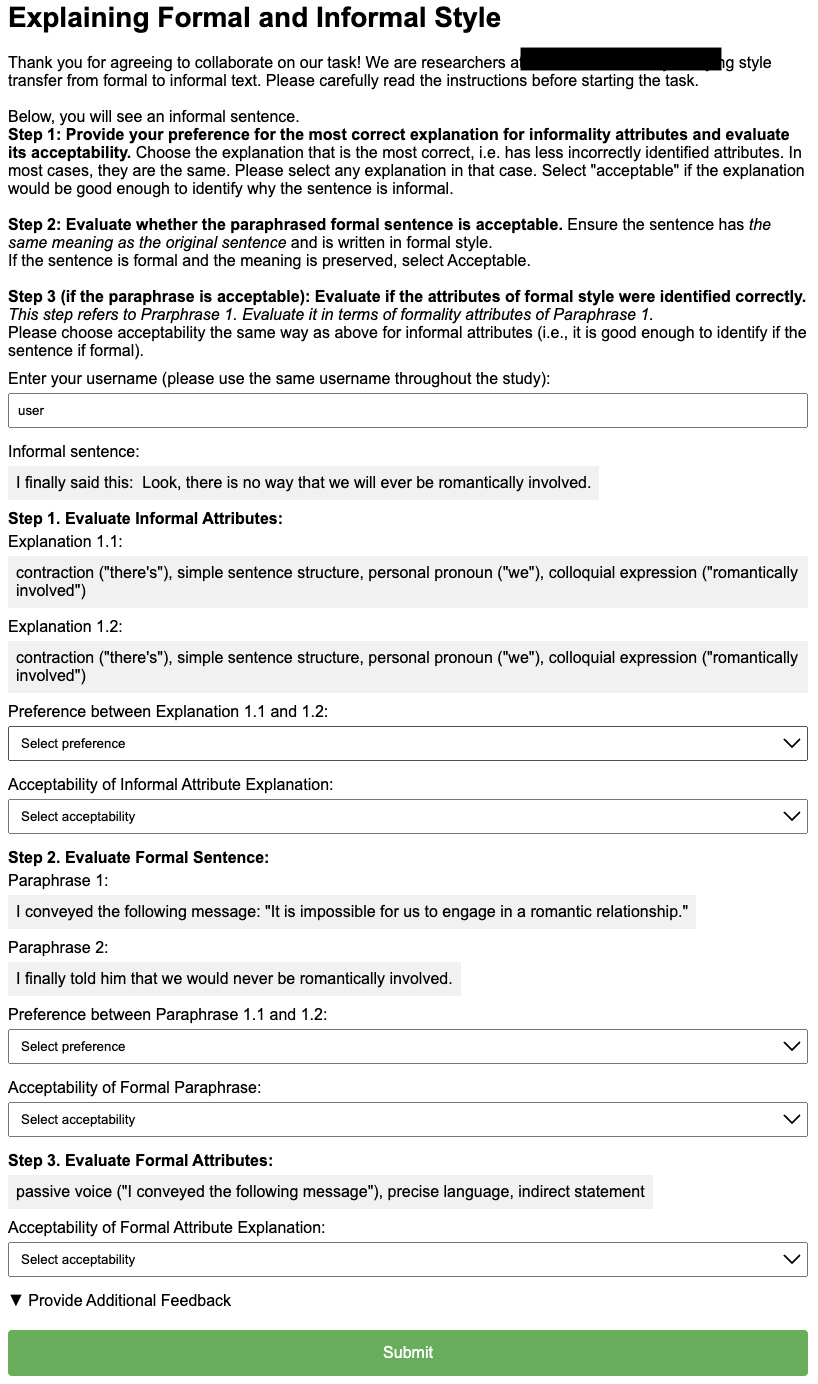}
    \caption{Annotation for eGYAFC data acceptability and preferences.}
    \label{fig:egyafc-annot}
\end{figure*}

\section{Annotator demographics} \label{app:demographics}
Annotators are part of a diverse demographic, geographically present in North America, Europe, and Southeast Asia (as reported by Upwork). All annotators indicated at least fluent to native English skill.

\section{AI assistants} \label{app:aiassist}
The authors used AI assistants such as Co-Pilot and ChatGPT for writing code.

\end{document}